\documentclass[11pt,a4paper]{article}
\usepackage[hyperref]{acl2020}
\usepackage{times}
\usepackage{latexsym}

\usepackage{soul}
\usepackage{url}
\usepackage{algorithm}
\usepackage{algorithmicx}
\usepackage[noend]{algpseudocode}
\usepackage{balance}
\newcommand{\ModelName}{\textsc{CG-BERT}}
\usepackage{amsmath,amsthm,mathtools}
\usepackage{latexsym,graphicx,amssymb,multirow,booktabs}
\usepackage{makecell}

\usepackage{subfigure}
\usepackage[font={footnotesize}]{caption}

\usepackage{hyperref}
\usepackage{url}
\usepackage{xcolor}
\usepackage{soul}
\usepackage{enumitem}
\usepackage{mathtools}
\usepackage{microtype}

\aclfinalcopy 
\def\aclpaperid{1818} 

\title{CG-BERT: Conditional Text Generation with BERT for Generalized Few-shot Intent Detection}

\author{}
\author{Congying Xia{$^{1}$},~Chenwei Zhang{$^2$}, Hoang Nguyen{$^{1}$}, Jiawei Zhang{$^{3}$}, \textbf{Philip Yu}{$^{1}$}\\
   {$^1$University of Illinois at Chicago, IL, USA;}  {$^2$Amazon, WA, USA;} \\
  {$^3$Florida State University, FL, USA}\\
  {\tt \{cxia8,hnguy7,psyu\}@uic.edu; }\\
  {\tt cwzhang@amazon.com; jzhang@ifmlab.org}
}
\date{}

\begin{document}
\maketitle
\begin{abstract}
In this paper, we formulate a more realistic and difficult problem setup for the intent detection task in natural language understanding, namely Generalized Few-Shot Intent Detection (GFSID). GFSID aims to discriminate a joint label space consisting of both existing intents which have enough labeled data and novel intents which only have a few examples for each class. To approach this problem, we propose a novel model, Conditional Text Generation with BERT ({\ModelName}). {\ModelName} effectively leverages a large pre-trained language model to generate text conditioned on the intent label. By modeling the utterance distribution with variational inference, CG-BERT can generate diverse utterances for the novel intents even with only a few utterances available.
Experimental results show that {\ModelName} achieves state-of-the-art performance on the GFSID task with 1-shot and 5-shot settings on two real-world datasets.
\end{abstract}

\section{Introduction}
Intent Detection (ID) is a key task in spoken dialogue systems, which aims at understanding the intents behind users' inputs \cite{tur2011spoken}. 
In real-world applications, such as Google Assistant, detecting the intents in users' utterances is crucial for downstream tasks, such as dialogue management and dialog state tracking.

In this ever-changing digital world, intelligent assistants need to have the ability to adapt to customers' new requests promptly. State-of-the-art intent detection models \citep{haihong2019novel, zhang2018joint, goo-etal-2018-slot} capitalize on large amounts of labeled data to train supervised deep learning classification models.
These models can be trained solely based on the existing intents, but they generalize poorly to novel intents. It is difficult and labor-intensive to collect large-scale, high-quality annotations on novel intents and re-train the whole model.

Inspired by human's ability to adapt existing knowledge to new concepts quickly with only a few examples, few-shot learning (FSL) \cite{fei2006one} has drawn a lot of attention recently.
FSL approaches \cite{vinyals2016matching} are expected to discriminate the novel classes from each other with only a few examples, namely, few-shots.
However, this formulation offers no incentive to maintain a globally consistent label space with the existing classes.
From a practical point of view, we would like the model to incorporate these novel classes which only have a few labeled examples into the label space of existing classes that have enough labeled data. This ability is crucial for the intent detection task in practice, as it relaxes the assumption that we only need to detect the novel intents.

\begin{figure*}[!ht]
    \centering
    \includegraphics[width=\linewidth]{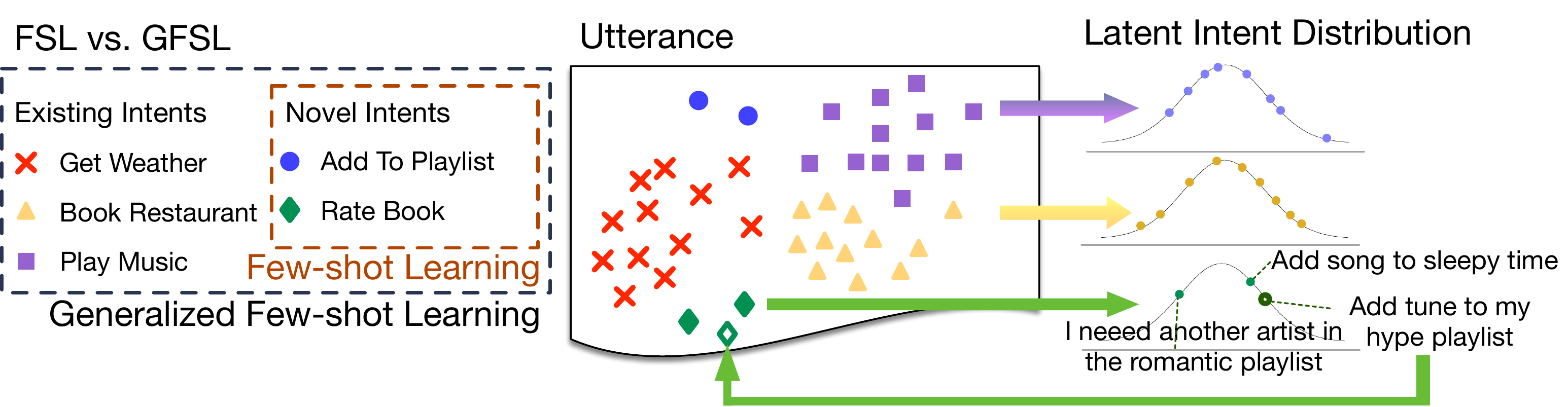}
    \vspace{-0.1in}
    \caption{Compared to FSL which only needs to discriminate among the novel intents, GFSID needs to discriminate all the intents including the existing ones and the novel ones. We model the distribution for the diverse utterances associated with one intent and generate new utterances by sampling from this learned distribution. We augmented the training dataset with the generated examples for the novel intents, and convert the task into a supervised classification task. 
    }
    \label{fig:overview}
\end{figure*}

In this work, we propose a more difficult yet realistic problem setup for intent detection in the low-resource scenarios, which is called Generalized Few-Shot Intent Detection (GFSID). As shown in the left part of Figure \ref{fig:overview}, GFSID aims to correctly classify utterances that belong to both existing and novel intents. 
This terminology is adopted from the Generalized Few-shot Learning \cite{xiahan19relation} which is an extension setup from FSL. The performance of FSL models deteriorates drastically in discriminating the joint label space, despite their good performance on novel label space \cite{xiahan19relation}. Compared to FSL, GFSID is a much more challenging task since the model has a bias on the existing classes over the novel classes and prefers to predict the test samples as the existing classes.

We propose to ease the scarce annotation problem in the GFSID task by generating more utterances for the novel intents and convert the generalized few-shot problem into a supervised classification problem.
As illustrated in Figure \ref{fig:overview}, we model the utterance distributions associated with the intents and generate new utterances for novel intents through sampling from this learned distribution.
However, it's difficult to learn a good distribution with only a few examples. To transfer the knowledge in a large amount of unlabeled data, we utilize the powerful pre-trained language model BERT \cite{devlin2018bert} to learn such a distribution. Since the distribution learned through BERT is unregulated for different intents, we propose to adopt the idea from Conditional Variational Auto-Encoder (CVAE) \cite{NIPS2014_5352} to add a latent space mapping for BERT and regularize the BERT feature space to form a unit Gaussian distribution conditioned on a certain intent.

In this paper, we propose a conditional text generation model,  Conditional Text Generation with BERT ({\ModelName}), which incorporates CVAE in BERT to model a distribution over diverse utterances having the same intent. 
Through the proposed model, we are able to generate more utterances for the novel intents and augment the training dataset.
We utilize the state-of-the-art text classification model BERT to do intent detection with the augmented dataset in a supervised way.

To summarize, our main contributions are:
\begin{itemize}[leftmargin=*]
    \setlength\itemsep{0mm}
     \item We propose a more realistic and challenging problem setup in low-resource conditions, Generalized Few-Shot Intent Detection (GFSID), which aims to discriminate a joint label space consisting of both existing intents which have enough labeled data and novel intents which only have a few examples for each class.
        
    \item A novel conditional text generation model, {\ModelName}, is proposed to solve the GFSID task. {\ModelName} uses a latent variable to model the probability distribution of diversely expressed utterances that belongs to a certain intent. It is able to generate pseudo labeled examples for the novel intents and alleviate the scarce annotation problem for GFSID.

    \item The experiments conducted for the proposed GFSID task show the effectiveness of our proposed model on two real-word intent detection datasets.
\end{itemize}
\vspace{-0.15in}
\section{Problem Formulation}
\vspace{-0.05in}
In this section, we give the definition for the generalized few-shot intent detection (GFSID) task. Given an existing intent set ${\mathcal Y_{ex}}$, a large amount of labeled examples $\mathcal D_{ex} = \{(x_i, y_i), i = 1, 2, ..., |\mathcal{D}_{ex}|\}$ with $y_i \in {\mathcal Y_{ex}}$ are available. 
The novel intent set can be denoted as $\mathcal{Y}_{novel}$, where each intent has $K$ examples $\mathcal D_{novel} = \bigcup_{n}\{(x_{n,k}, y_n)\}_{k=1}^K$. These two intent sets are disjointed, i.e., $\mathcal Y_{ex} {\cap}  \mathcal Y_{novel}=\varnothing$; the overall intent space can be denoted as $\mathcal Y_{joint} = \mathcal Y_{ex} \cup \mathcal Y_{novel}$.

Generalized few-shot intent detection intends to classify a given utterance not only as one of the existing intents but also as the novel intents. Formally, given a new query utterance $x$, the GFSID task aims at inferring the most likely intent of $x$, i.e., 
\vspace{-0.1in}
\begin{equation}
    \hat{y} = \mathop {\arg \max }\limits_{y \in {\mathcal Y_{joint}}} p\left( {y|x, \mathcal D_{ex}, \mathcal D_{novel}} \right).
    \vspace{-0.05in}
\end{equation}
Compared to the traditional few-shot classification task which only needs to separate the few-shot classes $\mathcal Y_{novel}$, GFSID is much more technically challenging and requires the discrimination of a much larger label space $\mathcal Y_{joint}$ instead.
\vspace{-0.05in}
\section{The Proposed Model}
\vspace{-0.05in}
\textbf{C}onditional Text \textbf{G}eneration with \textbf{BERT} ({\ModelName}) is proposed to model the distribution of diverse utterances with a given intent. It is able to generate more utterances for the novel intent through sampling from the learned distribution.

\begin{figure*}
    \centering
    \includegraphics[width=.88\linewidth]{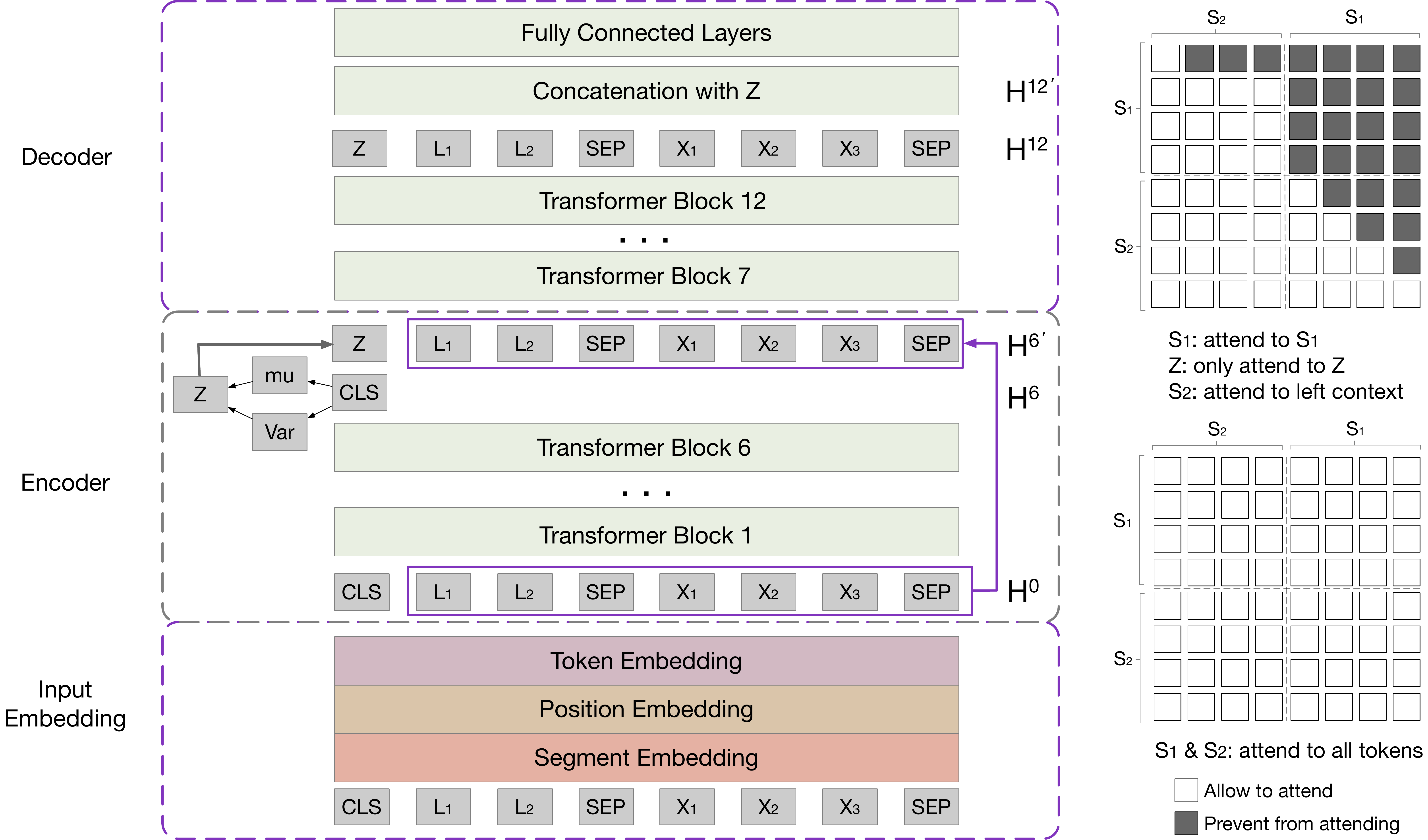}
    \caption{The overall framework of {\ModelName}: Conditional Text Generation with BERT. }
    \label{fig:proposed_model}
\end{figure*}

The overall framework of {\ModelName} is illustrated in Figure \ref{fig:proposed_model}. Particularly, {\ModelName} adopts the CVAE framework and incorporates BERT into both the encoder and the decoder. Formally, the encoder encodes the utterance $x$ and its intent $y \in \mathcal Y_{joint}$ together into a latent variable $z$ and models the posterior distribution $p(z|x, y)$, where $y$ is the condition in the CVAE model. The decoder decodes $z$ and the intent $y$ together to reconstruct the input utterance $x$.
To generate new utterances for an novel intent  $y \in \mathcal Y_{novel} $, we sample the latent variable $z$ from a prior distribution $p(z|y)$ and utilize the decoder to decode $z$ and $y \in \mathcal Y_{novel} $ into new utterances.

\subsection{Input Representation}
The input representation follows that of BERT \cite{devlin2018bert}.
In order to encode both the intent and the utterance, the input in our model is a pair of intent and utterance text sentences packed together.

Given an intent $y = (L_1, L_2, ..., L_{T_{1}}) $ with $T_1$ words and an utterance $x = (X_1, X_2, ..., X_{T_2})$ with $T_2$ words, we add special start-of-sequence ([CLS]) token at the beginning of the input and a special end-of-sequence ([SEP]) token at the end of each sentence. The first intent sentence, ([CLS], $L_1$, $L_2$, ...,  $L_{T_{1}}$, [SEP]), is referred to as $S_1$ and the second utterance sentence, ($X_1$, $X_2$, ..., $X_{T_2}$, [SEP]) is named as $S_2$. As shown in Figure \ref{fig:proposed_model}, $S_1$ and $S_2$ are concatenated together as the whole input. [SEP] not only marks the sentence boundary, but also is used for the model to learn when to terminate the decoding process. [CLS] is used as the representation for the whole input and we encode the embeddings for [CLS] to the latent variable $z$. 

Texts are tokenized into subword units by WordPiece \cite{wu2016google}. As illustrated in Figure \ref{fig:proposed_model}, three types of embedddings will be obtained for each token: token embeddings, position embeddings \cite{vaswani2017attention}, and segment embeddings \cite{devlin2018bert} which identifies the intent and the utterance. The input representation of a given token is constructed by summing these three embeddings and represented as $\mathbf{H}^0= [\mathbf{h}^{0}_1, ..., \mathbf{h}^{0}_{T}]$ with a total length of $T$ tokens.

\subsection{The Encoder}
As shown in Figure \ref{fig:proposed_model}, the encoder in our model encodes the input (both the intent and the utterance) into a latent variable that models the distribution of diverse utterances for a given intent.
To obtain deep bidirectional context information, we utilize multiple bidirectional Transformer encoders \cite{vaswani2017attention}, namely Transformer blocks in this paper, to encode the input. These Transformer blocks not only extract the semantic information for the input, but also models the attention between the intent tokens and the utterance tokens.

The BERT-base model is used as the backbone of {\ModelName} to incorporate the information pre-trained from a large amount of unlabeled data. As BERT-base has 12 transformer blocks, we utilize the first 6 transformer blocks in the encoder and the last 6 blocks are involved in the decoder.

The input representation $\mathbf{H}^0= [\mathbf{h}^0_1, ..., \mathbf{h}^0_{T}]$ is encoded into contextual layer representations through the transformer blocks. For each layer, the layer representation $\mathbf{H}^l= [\mathbf{h}^l_1, ..., \mathbf{h}^l_{T}]$ is computed with an $l$-layer Transformer block $\mathbf{H}^{l}=$ Transformer$_{l}(\mathbf{H}^{l-1}$), $l\in \{1, 2, \cdots, 6\}$. In each Transformer block, multiple self-attention heads are used to aggregate the output vectors of the previous layer. For the $l$-th Transformer layer, the output of a self-attention head $\mathbf{A}_l$ is computed via:
\vspace{-0.08in}
\begin{align}
\begin{aligned}
    \mathbf{Q} &= \mathbf{H}^{l-1}\mathbf{W}^l_Q, \\
    \mathbf{K} &= \mathbf{H}^{l-1}\mathbf{W}^l_K, \\
    \mathbf{V} &= \mathbf{H}^{l-1}\mathbf{W}^l_V, 
    \end{aligned}
\end{align}
\vspace{-0.1in}
\begin{equation}
\vspace{-0.08in}
{\mathbf{A}_l} = \text{softmax} \left(\frac{\mathbf{Q}\mathbf{K}^{\top}}{\sqrt {d_k}}\right){\mathbf{V}}\label{eq:att},
\end{equation}
where the output of the previous layer $\mathbf{H}^{l-1}{\in}\mathbb{R}^{T \times d_h}$
is linearly projected to a triple of queries, keys and values parameterized by matrices $\mathbf{W}^Q_l, \mathbf{W}^l_K, \mathbf{W}^l_V{\in}\mathbb{R}^{d_h{\times}d_k}$.
Following BERT, the embeddings for the [CLS] token in the 6-th transformer block $\mathbf{h}^6_1$ is regarded as the sentence-level representation. 

By modeling the true distribution $p(z|x, y)$ using a known distribution that is easy to sample from \cite{NIPS2014_5352}, we encode the sentence-level representation $\mathbf{h}^6_1$ into a latent variable $z$ whose prior distribution $p(z|y)$ is a multivariate standard Gaussian distribution.
We use the reparametrization trick \cite{kingma2013auto} to generate the latent vector $z$. The encoder predicts the parameters $\mu$ and $\sigma$ in the Gaussian distribution such that the proxy posterior $q(z|x, y) = \mathcal{N} (\mu, \sigma)$:

\begin{align}
	\begin{aligned}
	\mu  &= \mathbf{h}^6_1{\textbf{W}_\mu } + {b_\mu }, \\
    {\log(\sigma ^2)} &= \mathbf{h}^6_1{\textbf{W}_\sigma } + {b_\sigma }, \\
    {z} &= \mu + {\sigma}\varepsilon,
	\end{aligned}
\end{align}
where $\mu$ and $\log(\sigma ^2)$ are projected from $\mathbf{h}^6_1$ through parameters $\textbf{W}_\mu \in \mathbb{R}^{d_h \times d_h}$, $\textbf{W}_\sigma   \in \mathbb{R}^{d_h \times d_h}$, $b_\mu \in \mathbb{R}^{d_h}$, $b_\sigma \in \mathbb{R}^{d_h}$, $\varepsilon \in \mathcal{N}(0, 1)$ are used to sample $z$.

\subsection{The Decoder}
The decoder aims to reconstruct the input utterance $x$ using the latent variable $z$ and the intent $y$. As shown in Figure \ref{fig:proposed_model}, we add a residual connection from the input representation $\mathbf{H}^{0}$ to the input of the decoder $\mathbf{H}^{6'}$. The input of the decoder, $\mathbf{H}^{6'}= [z, \mathbf{h}^0_2, ..., \mathbf{h}^0_{T}]$, is constructed by concatenating the latent variable $z$ and the input representation $\mathbf{H}^{0}$ which represents the input tokens.
The input representation for the [CLS] token,  $\mathbf{h}^0_1$, is ignored since its position is taken by $z$.

The input for the decoder $\mathbf{H}^{6'}$ is fed into the last six transformer blocks of BERT-base, as shown in Figure \ref{fig:proposed_model}. To reconstruct the input utterance, the decoder is trained to generate the utterance in a left-to-right manner, since we can only use the left context to predict the next token. The original BERT model is trained on a large corpus to predict a randomly masked token given its contexts in the sentence. We try to leverage the conditional prediction ability and modify it for text generation in a left-to-right manner.

Inspired by the Unified Language model \cite{dong2019unified}, which utilizes specific self-attention masks to control what context the prediction conditions on, we adopt the attention mask which helps the transformer blocks fit into the conditional text generation task. Instead of applying the whole bidirectional attention to the input, a mask matrix $\mathbf{M} \in \mathbb{R}^{T \times T}$ is added to determine whether a pair of tokens can be attended to each other.
We update the calculation for the attentions in transformer blocks from Equation \ref{eq:att} to:

\vspace{-0.15in}
\begin{equation}
{\mathbf{A}_l} = \text{softmax} \left(\frac{\mathbf{Q}\mathbf{K}^{\top}}{\sqrt {d_k}}  + \mathbf{M} \right){\mathbf{V}}, 
\end{equation}
\vspace{-0.15in}
where \begin{equation}
\mathbf{M}_{ij} =
    \begin{cases}
      0, & \text{allow to attend;}\\
      -\infty,  & \text{prevent from attending.}
    \end{cases}
\vspace{-0.05in}
\end{equation}

The proposed attention mask matrix for the transformer blocks in the decoder is shown in the upper right corner of Figure \ref{fig:proposed_model}. To keep the transformers in the encoder having the same structure, we apply the bidirectional self-attention mask for the encoder as shown in the lower right corner of Figure \ref{fig:proposed_model}. The bidirectional self-attention mask allows all the tokens in $S_1$ and $S_2$ to attend to all the tokens.

In the attention mask for the decoder, the tokens in the first intent sentence $S_1$ can only attend to all the tokens in $S_1$, while the tokens in the second sentence $S_2$ can attend to both $S_1$ and all the left tokens in $S_2$. With this attention matrix, we are able to control the decoder to generate the output from left to right conditioned on the given intent.

A special token here is the first token which holds the latent variable $z$, it is only allowed to attend to itself.
This is due to the vanishing latent variable problem \cite{zhao2017learning} when adapting VAE/CVAE to natural language generators. Previous works \cite{zhao2017learning} reported the problem that the LSTM decoder tends to ignore the latent variable. We met the same issue when applying the variational autoencoder with transformers. To alleviate this problem, we only allow the latent variable $z$ to attend to itself to avoid it been overwhelmed by the information of other tokens. 

The output of the 12-th transformer block in the decoder is represented as $\mathbf{H}^{12} = [\mathbf{h}^{12}_1, \mathbf{h}^{12}_2, ..., \mathbf{h}^{12}_{T}]$, where $\mathbf{h}^{12}_1$ is the embeddings for the latent variable $z$.
To further increase the impact of $z$ and alleviate the vanishing latent variable problem, we concatenate the embedings of $z$ with all the tokens:
$\mathbf{H}^{12'} = [\mathbf{h}^{12}_1 \mathbin\Vert \mathbf{h}^{12}_1, \mathbf{h}^{12}_2 \mathbin\Vert \mathbf{h}^{12}_1, ..., \mathbf{h}^{12}_{T} \mathbin\Vert \mathbf{h}^{12}_1]$. 
Two fully-connected layers with a layer normalization \cite{ba2016layer} are applied to get the final representation:
\vspace{-0.1in}
\begin{equation}
    \mathbf{H}^{f} = g(f(f(\mathbf{H}^{12'}\mathbf{W_1} + b_1)\mathbf{W_2} + b_2)),
    \vspace{-0.1in}
\end{equation}
where $\mathbf{W}_1 \in \mathbb{R}^{2 d_h \times d_h}$,  $\mathbf{W}_2 \in \mathbb{R}^{d_h}$, $f$ is a Gelu \cite{hendrycks2016bridging} activation function and $g$ is for layer normalization. The embeddings in $\mathbf{H}^{f}$ at position at position $t$ is used to predict the next token at position $t+1$:
\vspace{-0.1in}
\begin{equation}
    p(X_{t+1}) = f(\mathbf{H}^{f}_t \mathbf{W_e}^{\top} + b_e),
    \vspace{-0.1in}
\end{equation}
where $\mathbf{W_e} {\in} \mathbb{R}^{V{\times}d_h }$ is the token embedding in the input representation, $b_e {\in} \mathbb{R}^{V}$, $V$ is vocabulary size.

\vspace{-0.01in}
\subsection{Loss Function}
We train this model to maximize $\log p(x|y)$, the conditional log-likelihood of the utterance $x$ given an intent $y$, which involves a marginalization over the latent variable $z$. As proposed in \cite{NIPS2014_5352}, CVAE can be effectively trained with the Stochastic Gradient Variational Bayes framework by maximizing the evidence lower bound (ELBO) of the log-likelihood. The loss function for the proposed model is:
\vspace{-0.1in}
\begin{align}
	\begin{aligned}
   \mathcal{L} =&  - 
    KL \left[ {q\left( {z|x,y} \right), p\left( {z|y} \right)} \right] + \\
    &{E_{q(z|x,y)}}\left[ {\log p\left( {x|z,y} \right)} \right],
    \vspace{-0.1in}
\end{aligned}
\end{align}
which consists of the KL divergence loss and the reconstruction loss.       
\vspace{-0.1in}
\subsection{Generating Utterances for Novel Intents}
The distributions of the intents are learned after training the {\ModelName}. Through sampling from these distributions, we are able to generate utterances for novel intents. As shown in Figure \ref{fig:proposed_model}, we use the modules in the purple boxes (Input Embedding and Decoder) to generate new utterances. For a given novel intent $y$, an intent sentence $S_1$ is fed into the Input Embedding module to get the input representation $\mathbf{H}^0$. The input for the decoder is constructed through $\mathbf{H}^{6'} =  [z, \mathbf{h}^{0}_2, ..., \mathbf{h}^{0}_{T_1}]$, where $z$ is sampled from a multivariate standard Gaussian distribution instead of encoded from the input. With the decoder, the new utterance is generated sequentially from left to right.

To generate more utterances, we sample $z$ for $s$ times, and for each $z$, we get the top 20 results using beam search. To increase diversity, we will delete the generated sentences which already exist in the training data. The procedure for generating new utterances for the novel intents is illustrated as:

\begin{algorithm}[h]
\caption{Text Generation for Novel Intents}
\label{al:algorithm}
\begin{algorithmic}[1]
\Procedure{Text Generation}{$y$, $s$, $k$}
    \For {$y$~all novel intents} 
       \For {$s$~sample\_times}
        \State {$z_s \in \mathcal{N}(0, 1)$}
     \State {$\mathcal{G}_y.\text{add}(\text{top}  k(decode(z_s, y)))$} \Comment{$\mathcal{G}_y$ is an utterance set with intent $y$.}
     \EndFor
     \State{$\mathcal{G}_y$ = $\mathcal{G}_y$.remove($\mathcal D_{novel}$)} \Comment{Remove the provided few-shot examples in $\mathcal{G}_y$.}
     \State{$\mathcal D_{ge}$.add(y, $\mathcal{G}_y$)} \Comment{$\mathcal D_{ge}$ is the generated dataset.}
     \EndFor
    \State Return $\mathcal D_{ge}$
\EndProcedure
\end{algorithmic}
\end{algorithm}
\vspace{-0.2in}

The generated dataset $\mathcal D_{ge}$ are merged with the original training data $\mathcal D_{seen}, \mathcal D_{novel}$ to do intent detection with fine-tuning on BERT.
\vspace{-0.05in}
\section{Experiments}
\vspace{-0.05in}
\subsection{Datasets}
To demonstrate the effectiveness of our proposed model, we evaluate {\ModelName} on two real-word datasets for the generalized few-shot intent detection task: SNIPS-NLU \cite{coucke2018snips} and NLU-Evaluation-Data (NLUED) \cite{XLiu.etal:IWSDS2019}. These two datasets were collected to benchmark the performance of natural language understanding services offering customized solutions. The statistical information of these two datasets is shown in Table \ref{dataset}.
\begin{table*}[ht!]
\centering
\resizebox{\linewidth}{!}{
\begin{tabular}{l|cccc|cccc}
\Xhline{3\arrayrulewidth}
 & Overall & Seen  & Novel & H-Mean & Overall & Seen  & Novel & H-Mean \\ 
                     & \multicolumn{4}{c|}{SNIPS-NLU 1-shot}     & \multicolumn{4}{c}{SNIPS-NLU 5-shot}    \\ \Xhline{3\arrayrulewidth}
IN+  &18.02 $\pm$ 2.92&20.16 $\pm$ 4.20	&12.38 $\pm$ 8.34	&13.44 $\pm$ 7.15 & 28.91 $\pm$ 1.44 &30.34 $\pm$ 1.42&25.15 $\pm$ 3.49&	27.39 $\pm$ 2.21  \\
BERT-SMOTE &74.65 $\pm$ 0.80& 96.24 $\pm$ 0.35& 17.66 $\pm$ 2.71& 29.77 $\pm$ 3.94&  83.56 $\pm$ 1.42 & 95.84 $\pm$ 1.08& 51.13 $\pm$ 4.04 & 66.61 $\pm$ 3.54\\
BERT                 & 83.42 $\pm$ 1.18 & 98.20 $\pm$ 0.06& 44.42 $\pm$ 4.35 & 57.74 $\pm$ 7.50  & 93.80 $\pm$ 1.74 &98.34 $\pm$ 0.10& 81.82 $\pm$ 6.16 &89.22 $\pm$ 3.74  \\
SVAE & 83.65 $\pm$ 1.55 &\textbf{98.24} $\pm$ \textbf{0.09} & 45.15 $\pm$ 5.54& 61.67 $\pm$ 5.11&93.88 $\pm$ 1.11 & \textbf{98.34} $\pm$ \textbf{0.06} & 82.10 $\pm$ 4.06&	89.49 $\pm$ 2.47  \\
BERT-PN+ &83.83 $\pm$ 4.13&92.66 $\pm$ 4.49 & 60.52 $\pm$ 7.58 &	72.99 $\pm$ 5.97 &93.23 $\pm$ 0.94 &	95.96 $\pm$ 1.13&	86.03 $\pm$	2.00&	90.71 $\pm$ 1.19  \\ 
{\ModelName}            & \textbf{88.49 $\pm$ 1.55}	
& 98.13 $\pm$ 0.15 & \textbf{63.04 $\pm$ 5.49} & \textbf{76.65 $\pm$ 4.24} & \textbf{95.16 $\pm$ 1.12}	
& 98.30 $\pm $0.17& \textbf{86.89 $\pm$ 4.05} & \textbf{92.20 $\pm$ 2.32}\\ 
\Xhline{3\arrayrulewidth}
                     & \multicolumn{4}{c|}{NLUED 1-shot}     & \multicolumn{4}{c}{NLUED 5-shot}  
                      \\ \Xhline{3\arrayrulewidth}
BERT-SMOTE  &52.44 $\pm$ 0.96 &85.89 $\pm$ 1.26 & 3.35 $\pm$ 1.37& 6.40 $\pm$ 2.51&64.17 $\pm$ 1.25 & 86.03 $\pm$ 1.32&32.11 $\pm$ 1.89 &46.74 $\pm$ 2.06\\
BERT                 & 72.11 $\pm$ 1.27
& 94.00 $\pm$ 0.93 & 7.88 $\pm$ 3.28 & 14.39 $\pm$ 5.66  &83.33
$\pm$ 0.74&\textbf{94.12 $\pm$ 0.89}& 51.69 $\pm$ 3.19 &	66.67 $\pm$ 2.51  \\
SVAE & 72.22 $\pm$ 1.07 & 93.80 $\pm$ 	0.70&  8.88 $\pm$ 3.66& 16.01 $\pm$ 6.06& 83.54 $\pm$ 0.88 & 93.60 $\pm$ 	0.63&	54.03 $\pm$ 3.91 &  68.42 $\pm$ 3.06 \\
BERT-PN+ & 71.49 $\pm$ 2.40&	81.24 $\pm$ 2.76&	18.95 $\pm$ 4.42&	30.67 $\pm$ 5.53 & 78.29 $\pm$ 2.62	 &83.41 $\pm$ 2.62&	60.28 $\pm$ 4.19&	69.93 $\pm$ 3.49
			\\
{\ModelName}  & \textbf{75.29 $\pm$ 1.75} &\textbf{94.01 $\pm$ 0.70} & \textbf{20.39 $\pm$ 5.77} & \textbf{33.12 $\pm$ 7.92} 
& \textbf{85.48 $\pm$ 0.88} & 93.80 $\pm$ 0.60 & \textbf{61.06 $\pm$ 4.29} & \textbf{73.88 $\pm$ 3.10}   \\\Xhline{3\arrayrulewidth}
\end{tabular}
}\vspace{-0.1in}
\caption{Generalized few shot experiments with 1-shot/5-shot setting on SNIPS-NLU and NLUED.}
\label{exp}
\end{table*}

\noindent\textbf{SNIPS-NLU\footnote{https://github.com/snipsco/nlu-benchmark/}} Following \citet{xia2018zero}, we select two intents (RateBook and AddToPlaylist) as emerging intents (the few-shot classes), while the other five intents are regarded as existing intents. For the GFSID task, we need to do classification among the whole 7 intents. We randomly choose 80\% of the whole data as the training data and 20\% as the test data. For the few-shot intents, we randomly sample 1 or 5 examples as the few-shots.

\noindent\textbf{NLUED} Following \citet{XLiu.etal:IWSDS2019}, a sub-corpus of 11, 036 utterances covering all the 64 intents are used. 
We randomly choose 16 intents as the few-shot ones. We perform the generalized intent detection task among the whole 64 intents over 10-Folds\footnote{https://github.com/xliuhw/NLU-Evaluation-Data}.

\begin{table}[ht!]
\centering
\resizebox{\linewidth}{!}{
\begin{tabular}{l|c|c}
\Xhline{3\arrayrulewidth}
 Dataset & SNIPS-NLU & NLUED \\ \hline
Vocab Size & 10,896 & 6,761\\ 
\#Total Classes & 7 & 64 \\ 
\#Few-shot Classes & 2 & 16\\
\#Few-shots / Class & 1 or 5 & 1 or 5 \\
\#Training Examples & 7,858 & 7,430\\ 
\#Training Examples / Class & 1571.6 & 155\\
\#Test Examples & 2,799 & 1,076\\ 
Average Sentence Length   &  9.05    & 7.68\\
\Xhline{3\arrayrulewidth}
\end{tabular}
}
\caption{Data Statistics for SNIPS-NLU and NLUED. \#Few-shot examples are excluded in the \#Training Examples. For NLUED, we report the statistics in KFold\_1.}
\label{dataset}
\end{table}
\vspace{-0.3in}

\subsection{Baselines}
To the best of our knowledge, we are the first one to study the GFSID task. Following \citet{xiahan19relation}, we extend two state-of-the-art FSL methods to the GFSID setup as our baselines.
From the data augmentation perspective, our model is an over-sampling model which samples sentence level examples from latent distributions. To show the effectiveness of our proposed model, we also compare it with other over-sampling methods. 

Five baselines are considered in total:
1) IN$^{+}$. Induction Networks \cite{geng2019induction} (IN) is a state-of-the-art few-shot text classification model. It can take any-way any-shot inputs, hence we can easily apply a readily trained IN model to solve the GFSL task. 
2) BERT-PN$^{+}$. Prototypical Networks \cite{snell2017prototypical} (PN) is a simple but effective FSL model. To provide a fair comparison, we use BERT as the encoder in PN and finetune BERT together with the PN model.
3) BERT. BERT \cite{devlin2018bert} is the state-of-the-art text classification model. We simply over sample the few-shots by copying the few-shot examples to the maximum training examples per class and use the over-sampled dataset to do intent detection with fine-tuning on BERT.
4) BERT-SMOTE. SMOTE \cite{chawla2002smote} is an over-sampling method that samples features based on the nearest neighbors. BERT-SMOTE uses fixed pre-trained BERT features. A fully connected layer is used for intent detection with the over-sampled features.
5) SVAE. SVAE \cite{bowman2015generating} is a text generation model that utilizes Variational Auto-Encoder (VAE) to model the distribution of the utterances. It can generate similar sentences by sampling from the posterior or get intermediate sentences between two sentences.

\subsection{Implementation Details}
We use BERT-base and Adam optimizer \citep{kingma2014adam} for all the experiments. For {\ModelName}, the hidden dimension $d_h$ is 768 and the batch size is 16. {\ModelName} is trained with learning rate equals 1e-5 in 100 epochs and each epoch has 1000 steps. We sample the latent variable $s=10$ times and choose the top $k=20$ utterances when generating new utterances. For fine-tuning on BERT, we set batch size as 32, learning rate as 2e-5 and the number of the training epochs as 3.
For the FSL baselines, we train these models in the episode training method. For SNIPS-NLU, we train these models in 3-way for 1000 episodes. For NLUED, we train these models in 10-way for 1500 episodes.

\vspace{-0.1in}
\subsection{Experiment Results}
The generalized few-shot intent detection results on two datasets with 1-shot/5-shot settings are reported in Table \ref{exp}. In addition to the overall accuracy (Overall) measured on all intents, we follow
the convention in GFZL \cite{xian2017zero} and report the accuracy on seen intents (Seen), the accuracy on novel intents (Novel) together with their harmonic mean (H-Mean). We report the average and the standard deviation for SNIPS-NLU over 5 runs. The results on NLUED are reported over 10 Folds.

On both two datasets with 1-shot and 5-shot settings, our proposed model {\ModelName} achieves state-of-the-art performance on overall accuracy, novel accuracy, and harmonic mean. The improvement mainly stems from the high quality of the generated examples for novel intents, which leads to significantly increased novel accuracy and harmonic mean. {\ModelName} achieves the best performance for the seen accuracy on the NLUED 1-shot setting, while obtaining comparable results with BERT on three other settings. 

Among the extended few-shot learning models, BERT-PN+ achieves decent performance on the novel accuracy compared to other baselines while it sacrifices the performance on the seen intents.
Compared to BERT-PN+, {\ModelName} achieves good performance consistently by training the model on both seen intents and novel intents altogether.
IN+ lacks the ability to extend from the FSL training to the GFSID setting since it uses a parametric method other than distance-based methods like PN to do classification. It performs poorly on the NLUED dataset ($<$ 20\% on the overall accuracy, thus not reported in Table \ref{exp}). 

For the text generation baseline, SVAE performs worse than {\ModelName} on the novel intents due to the generated sentences are too similar to the input few-shots.
BERT-SMOTE also does not perform well since it uses fixed BERT features and does over-sampling by exploring the interpolation among the few-shots in the feature space.

\vspace{-0.1in}
\subsection{Evaluation of the Generated Utterances}
\vspace{-0.05in}
We show some examples generated by {\ModelName} for two novel intents in Table \ref{case}. The generated utterances for Rate Book show that {\ModelName} is able to generate similar utterances to the few-shots by adding expressions like ``can you" or replacing words (``like'' and ``want''). The cases in Alarm Query are pretty diverse. It learns from other intents like ``Calendar Query'' that expressions like ``show me'' can be used for request information. It also generates utterances that are totally different from the few-shots, like ``do i have any alarms''.

\begin{table}[hbt!]
\centering
\resizebox{0.92\linewidth}{!}{
\begin{tabular}{l}
\Xhline{3\arrayrulewidth}
Rate Book (Real examples)\\ \hline
rate the current essay 0 \\
rate this textbook four stars\\ 
i d like to rate my beloved world two points \\
\Xhline{3\arrayrulewidth}
Rate Book (Generated Examples)\\\hline
\textcolor{red}{can you} rate the current essay 0 \\
rate \textcolor{red}{that} textbook four stars \\
i d \textcolor{red}{want} to rate my beloved world two points \\
\Xhline{3\arrayrulewidth}
Alarm Query (Real examples)\\ \hline
my alarms \\
what alarms do i have set \\
tell me what alarms are set for me \\
what are my alarms \\
 \Xhline{3\arrayrulewidth}
Alarm Query (Generated Examples)\\\hline
\textcolor{red}{show me} my alarms \\
what alarms do \textcolor{red}{you} have set \\
\textcolor{red}{i need to see what alarms are there}\\
\textcolor{red}{do i have any alarms} \\
\textcolor{red}{remind me} what alarms are set for me \\
\Xhline{3\arrayrulewidth}
\end{tabular}
}
\vspace{-0.1in}
\caption{Generated examples for two novel intents. Rate Book is an novel intent in SNIPS-NLU and Alarm Query is an novel intent in NLUED.}
\label{case}
\end{table}

There are also some bad cases when the intent is not fully conditioned as expected. For example, ``am i going to need a jacket tomorrow" which has the intent of Query Weather, but generated for Alarm Query. In other cases, sentences are not well generated and may have syntax errors, like ``alarms me are there any alarms for me".

We further evaluate the quality of the generated utterances both quantitatively and qualitatively. Quantitative metrics like the percentage of unique sentences and n-grams are used. Sentences are manually annotated for qualitative analysis. We use the Fold 7 of the NLUED dataset with the 5 shot setting as an example. There are 16 novel intents in total and each novel intent has 5 examples. 
We randomly sample the hidden variable z for once and get the top 20 generated sentences for each intent. After deduplication, we got 257 unique sentences out of 320, which is around 80\%. 

\begin{table}[ht!]
\begin{tabular}{l|ccc}
\Xhline{3\arrayrulewidth}
        & unigram & bigram  & trigram \\
SVAE    & 3.57\%  & 19.51\% & 22.90\% \\
{\ModelName} & 31.74\% & 43.29\% & 47.36\% \\\Xhline{3\arrayrulewidth}
\end{tabular}
\caption{The average percentage of new n-grams in the generated sentences.}
\label{evaluation}
\end{table}
Quantitatively, we compare the average percentage of new n-grams in the generated sentences to a text generation model SVAE \cite{bowman2015generating}. As shown in Table \ref{evaluation}, {\ModelName} is able to generate more new n-grams which present the diversity of the generated sentences.
Qualitatively, we asked human annotators to annotate the syntax quality (result: 4.3 out of 5) and intent coherency (result: 4.7 out of 5) of the generated utterances.

\subsection{Visualization}
\begin{figure}[hbt!]
    \centering
    \includegraphics[width=.85\linewidth]{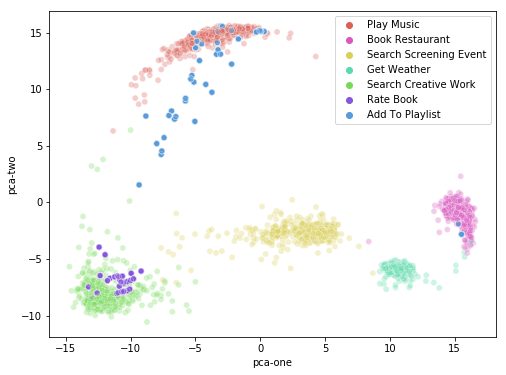}
    \vspace{-0.18in}
    \caption{t-SNE visualization of the embedding space for SNIPS-NLU with generated utterances from {\ModelName}.}
    \label{fig:latent_cg}
\end{figure}

\begin{figure}[hbt!]
    \centering
    \includegraphics[width=.85\linewidth]{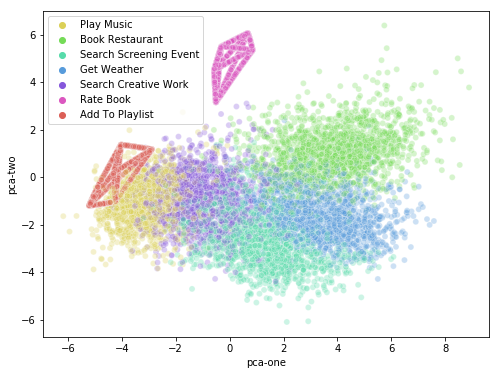}
    \vspace{-0.18in}
    \caption{t-SNE visualization of the embedding space for SNIPS-NLU with over-sampled features from BERT-SMOTE.}
    \label{fig:latent_smote}
\end{figure}

We visualize the embedding space of the generated utterances using t-SNE for BERT-SMOTE and CG-BERT. As shown in Figure \ref{fig:latent_cg}, the generated examples from {\ModelName} for novel intents are diverse and well separated from the seen intents, while BERT-SMOTE tends to generate similar examples on the feature level as in Figure \ref{fig:latent_smote}. SMOTE only generates new features within the few-shots, while {\ModelName} is able to generate diverse examples beyond these five shots by transfer expressions from existing intents. 

\vspace{-0.1in}
\section{Related Work}

\noindent\textbf{Few Shot Learning} \cite{miller2000learning, fei2006one} is a task that aims to learn classifiers for new classes with only a few training examples per class. Recent deep learning-based FSL approaches mainly fall into two categories:
(1) metric-based approaches, such as Matching Networks \cite{vinyals2016matching} and  Prototypical Networks \cite{snell2017prototypical}, which tries to learn one generalizable metric to separate the classes based on the distance for all the tasks; (2) optimization-based approaches, e.g. Meta Network \cite{munkhdalai2017meta} and MAML \cite{finn2017model}, which aims to optimize model parameters based the gradients computed from few-shot examples. 

Recently, some few-shot learning studies are presented with a special focus on few-shot text classification problems. 
\citet{rios-kavuluru-2018-shot} develop a few-shot text classification model for multi-label text classification where there is a known structure of the label space. 
\citet{geng2019induction} propose Induction Networks that use dynamic routing induction method to encapsulate the abstract class representation from a few examples. 
\citet{xu2019open} proposes an open-world learning model to deal with the unseen classes in the product classification problem. \citet{zhang2019improving} is the first work which utilizes the pre-trained language model for few-shot text classification task.

\noindent\textbf{Generalized FSL} extends the setup from FSL, where the model is required to perform classification on the joint label space consisting of both previously seen and novel classes. This setup is not yet well-studied, especially for the intent detection task. There are only several works in the field of computer vision which try to solve this problem. \citet{gidaris2018dynamic} utilizes an attention-based weight generator for novel classes to extend the classifier from seen classes to the joint label space. \citet{xiahan19relation} incorporates inter-class relations using graph convolution to embed novel class representations into the same space with seen classes. \citet{schonfeld2019generalized} utilizes Variational Autoencoders to align the distributions learned from images and use side-information to construct latent features that contain the essential multi-modal information associated with unseen classes.
\citet{ye2019learning} proposes a learning framework, Classifier Synthesis Learning (CASTLE), which learns how to synthesize calibrated few-shot classifiers in addition to the classifiers of seen classes, leveraging a shared neural dictionary between seen classes and novel classes.

\noindent\textbf{Conditional Text Generation.} As text generation is an important and difficult task in natural language processing, a lot of works have been presented to solve this task. Seq2seq models are standard encoder-decoder models widely used in text applications like machine translation \cite{luong2015effective}. Variational Auto-Encoder (VAE) models are another important family \cite{kingma2013auto} and they consist of an encoder that maps each sample to a latent representation and a decoder that generates samples from the latent space. The advantage of these models is the variational component and its potential to add diversity to the generated data.  They have been shown to work well for text generation \cite{bowman2015generating}. 

Conditional VAE (CVAE) \cite{NIPS2014_5352} is proposed to improve over seq2seq models for generating more diverse and relevant text. CVAE based models \cite{rajeswar2017adversarial, zhao2017learning} incorporate stochastic latent variables that represent the generated text.
\citet{malandrakis2019controlled} is the most relevant work, which investigates the use of text generation techniques to augment the training data for intelligent artificial agents. However, their generation method is template-based and they don't leverage the power of pre-trained language models \cite{devlin2018bert}. This is the first work that employs a text generation model to solve the generalized few-shot learning problem.
\vspace{-0.05in}
\section{Conclusions}
\vspace{-0.05in}
A new task named Generalized Few-Shot Intent Detection is studied in this paper. It aims at discriminating a joint label space of existing intents with enough annotation and novel intents with only a few examples. A novel model, Conditional Text Generation with BERT, is proposed to solve this task by generating new utterances conditioned on a given novel intent. To the best of our knowledge, this is the first work that alleviates the scarce annotation problem in intent detection by generating sentence level examples. The proposed model achieves state-of-the-art performance on two-real word intent detection datasets.

\bibliographystyle{acl_natbib}
\bibliography{acl2020}
\end{document}